\documentclass{Interspeech2024}

\interspeechcameraready

\usepackage{caption}
\usepackage{subcaption}
\usepackage{multirow}
\usepackage{enumitem}
\usepackage{graphicx}
\usepackage{booktabs}
\usepackage{tikz}

\usepackage{ulem}

\title{Towards scalable efficient on-device ASR with transfer learning}

\name{Laxmi}{Pandey}
\name{Ke}{Li}
\name{Jinxi}{Guo}
\name{Debjyoti}{Paul}
\name{Arthur}{Guo}
\name{Jay}{Mahadeokar}
\name{Xuedong}{Zhang}

\address{
  Meta, Menlo Park, CA
  }
\email{laxmip@meta.com}


\begin{document}

\maketitle

\begin{abstract}
    Multilingual pretraining for transfer learning significantly boosts the robustness of low-resource monolingual ASR models. This study systematically investigates three main aspects: (a) the impact of transfer learning on model performance during initial training or fine-tuning, (b) the influence of transfer learning across dataset domains and languages, and (c) the effect on rare-word recognition compared to non-rare words. Our finding suggests that RNNT-loss pretraining, followed by monolingual fine-tuning with Minimum Word Error Rate (MinWER) loss, consistently reduces Word Error Rates (WER) across languages like Italian and French. WER Reductions (WERR) reach 36.2\% and 42.8\% compared to monolingual baselines for MLS and in-house datasets. Out-of-domain pretraining leads to 28\% higher WERR than in-domain pretraining. Both rare and non-rare words benefit, with rare words showing greater improvements with out-of-domain pretraining, and non-rare words with in-domain pretraining.
\end{abstract}

%
%
%
%
%
\section{Introduction}

Automatic Speech Recognition (ASR) systems based on end-to-end neural networks face challenges when languages lack ample training data, resulting in suboptimal performance. Previous studies have explored various techniques for training ASR in low-resource languages \cite{thomas2012multilingual, kunze2017transfer, khare2021low}. One such strategy involves leveraging transfer learning.
Transfer learning (TL) leverages the knowledge gained from other high-resource languages to enhance the performance of ASR models in low-resource settings \cite{kunze2017transfer, ghoshal2013multilingual, huang2013cross, conneau2020unsupervised, huang2014multi}. 
Multilingual pretraining extends this principle of TL by using data from a wide range of languages to create a shared, multilingual representation capturing phonetic and linguistic characteristics. This shared representation is then fine-tuned for the target low-resource language, allowing the ASR system to leverage the phonetic and linguistic characteristics learned from the diverse set of languages during pretraining. As a result, the ASR system becomes more robust, adaptable, and accurate when it comes to transcribing speech in previously underrepresented languages.

\begin{figure}[ht]
\centering
\includegraphics[trim={0 1cm 0 0.8cm},width=0.48\textwidth]{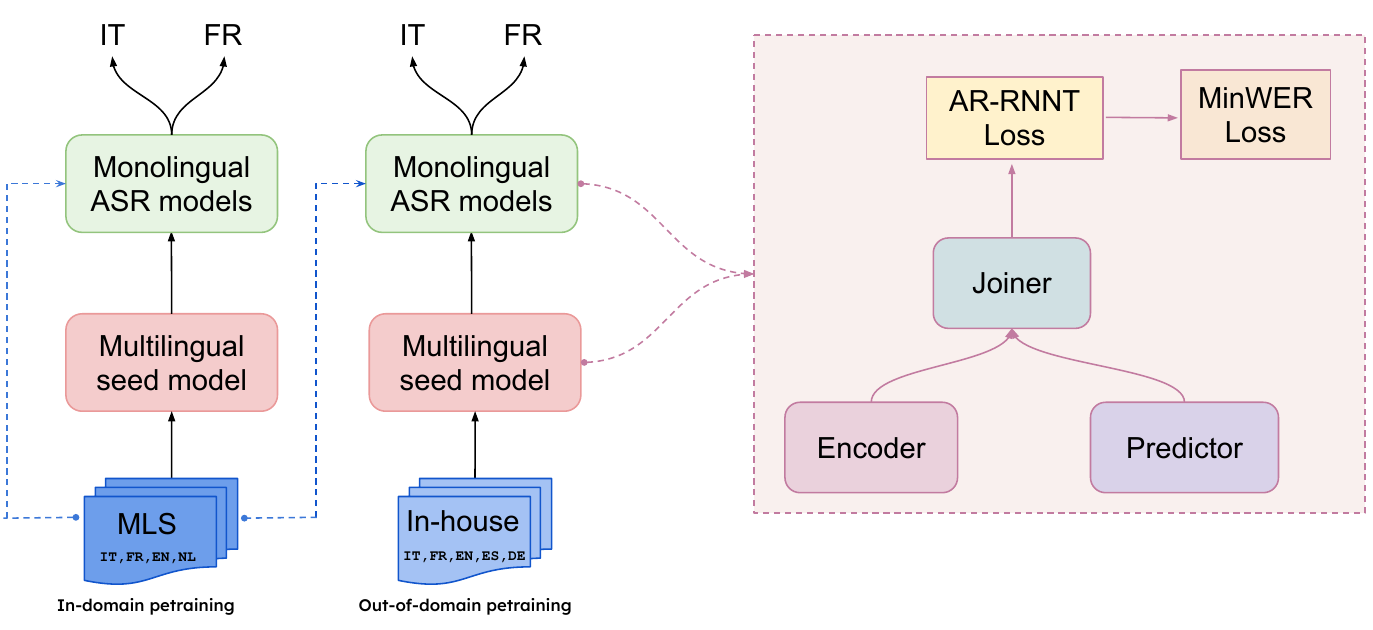}
\label{tl-block}
\caption{Visual representation of our multilingual pretraining strategy, showcasing both in-domain (MLS:seed | MLS:target) and out-of-domain (In-house:seed | MLS:target) approaches, alongside the ASR model training architecture with Alignment Restricted RNNT (AR-RNNT) and MinWER loss function.}
\end{figure}

Our research investigates the substantial enhancement in accuracy and efficiency achieved by employing multilingual transfer learning through pretraining in monolingual Automatic Speech Recognition (ASR) models for languages with limited training data. Our main objective is to explore the most effective approach for implementing pretraining-based transfer learning across languages, aiming to quantify its influence on both the quality of performance and computational efficiency. The innovation in our work does not solely rest on the transfer learning technique but rather on our comprehensive analysis of its effective utilization.

\begin{itemize}[leftmargin=*]
\item We first study the efficacy of transfer learning (TL) at two important stages of model training: initial training with the RNNT loss, and fine-tuning with the MinWER loss, with the goal to scrutinize whether TL exhibits greater effectiveness in one stage as opposed to the other.
\item We then explore the impact of domain-specific pretraining. We aim to investigate the comparative effectiveness of out-of-domain pretraining, utilizing multilingual data, against in-domain pretraining with multilingual data for the purpose of transfer learning.
\item Our third inquiry looks at whether transfer learning confers a more substantial benefit to rare words compared to common (non-rare) words. This specific investigation focuses on the impact of both in-domain and out-of-domain pretraining on the performance of rare and non-rare words.
\end{itemize}

\section{Related Work}
Several previous studies have endeavored to enhance low-resource languages by employing multilingual pretraining techniques, a technique that harnesses the knowledge gained from resource-rich languages to enhance ASR performance for those with limited training data.
The literature on multilingual pretraining for transfer learning in ASR underscores the promise of this approach in achieving cross-linguistic robustness \cite{ghoshal2013multilingual, huang2013cross, conneau2020unsupervised, babu2021xls}.

Various strategies have been proposed to enhance the performance of low-resource Automatic Speech Recognition (ASR) models \cite{thomas2012multilingual, swietojanski2012unsupervised, heigold2013multilingual, arsikere2019multi}, including transfer learning \cite{kunze2017transfer, huang2014multi}, multi-task training \cite{yu2012efficient, huang2013cross}, and ensemble learning \cite{deng2014ensemble, elfeky2016towards}. Transfer learning involves leveraging a well-trained Acoustic Model (AM) from a high-resource language to bootstrap the low-resource AM. Multi-task training and ensemble learning aim to utilize multilingual data and share model parameters, demonstrating successful strategies in low-resource scenarios. Recent advancements in the End-to-End (E2E) framework have introduced multilingual approaches, such as the multilingual RNNT model with language-specific adapters and datasampling to address data imbalance \cite{kannan2019large, thomas2022efficient}. Another work introduces an approach featuring an audio-to-byte End-to-End (E2E) system \cite{li2019bytes}, leveraging bytes as target units for enhanced scalability across multiple languages. Additionally, a transformer-based multilingual End-to-End (E2E) model has been introduced, integrating language information into the model at both the decoder and encoder levels. This involves the use of language identity tokens, and providing language information to acoustic vectors through one-hot vectors or learned language embeddings \cite{shetty2020improving}. While multilingual methods are appealing for low-resource languages, transfer learning stands out for its simplicity and effectiveness, requiring only pre-trained models without the need for the original high-resource language data. In our exploration of strategies to improve low-resource RNNT models, we focus on the simplicity and effectiveness of transfer learning.

\begin{table*}[h]
\begin{center}
\begin{tabular}{|l|ccc|ccc|}
\hline
\multirow{2}{*}{\textbf{Model}}    & \multicolumn{3}{c|}{\textbf{MLS}}                                                         & \multicolumn{3}{c|}{\textbf{In-house}}                                                  \\ \cline{2-7} 
                                   & \multicolumn{1}{c|}{\textbf{FR}}   & \multicolumn{1}{c|}{\textbf{IT}}    & \textbf{Average}   & \multicolumn{1}{c|}{\textbf{FR}}   & \multicolumn{1}{c|}{\textbf{IT}}   & \textbf{Avg}  \\ \hline
A. Monolingual ASR {\scriptsize{(baseline)}}                    & \multicolumn{1}{c|}{14.99}         & \multicolumn{1}{c|}{27.78}          & 21.38          & \multicolumn{1}{c|}{15.3}          & \multicolumn{1}{c|}{8.32}          & 11.80         \\ \hline
B. Multilingual ASR {\scriptsize{(seed)}}                   & \multicolumn{1}{c|}{18.93}         & \multicolumn{1}{c|}{38.14}          & 28.53~\scriptsize{(33.4\%$\uparrow$})         & \multicolumn{1}{c|}{21.35}         & \multicolumn{1}{c|}{14.59}         & 17.97~\scriptsize{(52.2\%$\uparrow$})        \\ \hline
C. B seeded monolingual RNNT ASR                  & \multicolumn{1}{c|}{11.01}         & \multicolumn{1}{c|}{19.73}          & 15.37~\scriptsize{(28.1\%$\downarrow$})          & \multicolumn{1}{c|}{11.4}          & \multicolumn{1}{c|}{4.31}          & 7.85~\scriptsize{(33.4\%$\downarrow$})          \\ \hline
D. B seeded monolingual MinWER ASR              & \multicolumn{1}{c|}{16.81}         & \multicolumn{1}{c|}{28.93}          & 22.87~\scriptsize{(6.9\%$\uparrow$})          & \multicolumn{1}{c|}{17.4}          & \multicolumn{1}{c|}{10.33}         & 13.85~\scriptsize{(13.3\%$\uparrow$})         \\ \hline
\textbf{E. C seeded monolingual MinWER ASR} & \multicolumn{1}{c|}{\textbf{9.71}} & \multicolumn{1}{c|}{\textbf{17.57}} & \textbf{13.64~\scriptsize{(36.2\%$\downarrow$})} & \multicolumn{1}{c|}{\textbf{9.61}} & \multicolumn{1}{c|}{\textbf{3.87}} & \textbf{6.74~\scriptsize{(42.8\%$\downarrow$})} \\ \hline
\end{tabular}
\end{center}
\caption{Effectiveness of Multilingual Encoder-Only Pretraining during RNNT Training and MinWER Finetuning Stages Across Languages: A Comparative Analysis Against Multilingual and Monolingual Baselines.}
\label{table1}
\end{table*}

\section{Model}
\subsection{AR-RNNT Loss Training}
An RNNT model comprises three main components: an encoder, a prediction network, and a joiner network. The encoder processes acoustic feature vectors $x_t$ into a sequence of hidden states $h^{enc}t$ over time $t$. The prediction network takes the previous sub-word label prediction $y{u-1}$ and produces a hidden representation $h^{pre}_u$ for label index $u$. The joiner network combines encoder output $h^{enc}t$ and prediction network output $h^{pre}u$ to compute output logits $h{t,u}$, followed by a softmax operation to obtain final posterior probabilities $P{t,u}$ for output tokens.

In the context of the AR-RNNT loss function, alignment constraints are imposed to ensure the alignment between input and output sequences follows specific restrictions, beneficial for tasks like speech recognition. The loss function of AR-RNNT is derived from the negative log posterior of the output label sequence $y$ given the input acoustic feature $x$, incorporating alignment constraints to enhance model performance in sequence-to-sequence tasks.
\begin{equation}
 L_{RNNT} = -log P(y|x)
\end{equation}

where $P(y|x) = \sum_{\hat{y}}P(\hat{y}|x), \hat{y}\in A$. A is the set of all
the possible alignments (containing both blank and non-blank
labels) between input x and output y.

\subsection{MinWER Loss Finetuning}
The MinWER-loss fine-tuning adopts the alignment-restricted MinWER loss and set-up proposed in \cite{guo2024-AR-MWER}, which is an extension of the original RNNT MinWER training method introduced in \cite{guo2020efficient}. This method is specifically engineered to prioritize training with the WER objective and enable discrinimative training. It does so by employing a loss function based on the weighted average of word errors within N-best hypotheses. 
\begin{equation}
    L_{MinWER} = \sum_{y_i} \hat{P}(y_i | x)R(y_i,y^r)
\end{equation}
Here, $\hat{P}(y_i | x)$ represents the posterior probability of a hypothesis $y_i$, while $R(·)$ quantifies the risk function, gauging the edit-distance at the word level between the hypothesis yi and the reference transcription $y^r$ in MinWER training.

\section{Experimental Setup}
\subsection{Training}
For the baseline RNNT model, the encoder is a 20-layer streamable low-latency Emformer with 40 ms lookahead, and process audio segments of 160 ms. It has input dimension of 512, hidden dimension 2048, 8 self-attention heads, and 1024-dimensional FC projection. The predictor consists of three LSTM layers with 512
hidden dimension. 
The output vocabulary comprises 5001 unigram SentencePieces \cite{kudo2018subword}, derived from the training transcripts, including the blank symbol We train all models using Alignment Restricted RNNT
loss \cite{mahadeokar2020alignment}, where the alignment is provided by a hybrid acoustic model.

Additionally, regarding the pretraining methodology, we explored different scenarios where we could apply pretraining to these components individually or collectively. Interestingly, our findings revealed that applying pretraining exclusively to the encoder component yielded better convergence compared to the alternative scenarios. This suggests that focusing pretraining efforts solely on the encoder enhances the model's ability to learn and adapt effectively to the given task. As a result, we have decided to adopt encoder-only pretraining consistently for all models in this study.

\subsection{Dataset}
\label{datasets}
We utilize two multilingual datasets for our experiments:
\begin{itemize}[leftmargin=*]
\item \textit{Multilingual Librispeech (MLS) dataset \cite{pratap2020mls}:} The MLS dataset contains multilingual speech derived from read audiobooks. We select four languages to be studied, English (EN), French (FR), Italian (IT), and Dutch (NL). The sizes of the training audio for the 4 languages are 44.7k hrs, 1.1k hrs, 0.2k hrs, 1.6k hrs, respectively. We categorized Italian (IT) and French (FR) as low-resource languages due to the relatively limited amount of training data and development hours allocated to these languages when compared to others. For evaluation purposes, we used the MLS test set.

\item \textit{In-house datasets:} This dataset is a collection of data from various sources. These datasets were collected from third-party vendors via crowd-sourced volunteers responding to artificial prompts with mobile devices. The content varies from voice assistant commands to simulations of conversations between people, as well as Facebook products such as public Facebook videos and voice commands to Portal. Videos used in this dataset are from 120 different countries. Additionally, data was generated from an in-house TTS model to increase the diversity of sentence patterns in our training data. All in-house datasets are de-identified. Depending on the source, the data was further augmented with various distortion methods, including speed perturbation, simulated reverberation, and randomly sampled additive background noise extracted from public Facebook videos. We selected five languages for in-domain datasets, including English (EN), French (FR), Italian (IT), German (DE), and Spanish (ES). The sizes of the training audio for the 4 languages are 1.6k hrs, 0.6k hrs, 0.5k hrs, 0.3k hrs, 0.9k hrs, respectively. In alignment with MLS's classification of low-resource languages, we have similarly categorized IT and FR as low-resource languages for this dataset. For evaluation purposes, we use handtranscribed data from the Ray Ban / Meta Smartglasses, Portal, Video, and Conversational data sources, ranging from 3K-15K utterances with no overlap with the training data. 

\end{itemize}

\section{Experiments}
\subsection{Impact of Transfer Learning: RNNT Training vs. MinWER Finetuning}
This study focuses on examining the influence of multilingual pretraining on monolingual ASR models. It investigates whether applying pretraining proves beneficial at the RNNT training stage or during Minwer finetuning.

We initially trained two multilingual models on the MLS dataset and the in-house dataset, respectively. These multilingual models served as the foundation for training monolingual models. We established a baseline by training monolingual ASR models on individual language data from MLS and in-house datasets. Subsequently, we conducted a performance analysis, comparing four distinct configurations. 
\begin{enumerate}

\item In the first configuration, we trained a multilingual model with combined data from various languages [section \ref{datasets}], using it as a seed model. 
\item The second configuration involved applying pretraining to the RNNT training stage. 
\item In the third configuration, pretraining took place during the MinWER finetuning stage. 
\item Lastly, our fourth configuration utilized pretraining during the RNNT training stage, followed by fine-tuning the converged model using MinWER optimization. 
\end{enumerate}

Our findings suggest that the third configuration works best, which involves initial pretraining during the RNNT stage followed by fine-tuning with MinWER. This configuration demonstrates a substantial improvement, achieving an average WERR of 36.2\% and 42.8\% compared to the baseline for the MLS and in-house datasets, respectively. We employ this best-performed configuration for all subsequent experiments in this study.

\subsection{Influence of Domain-Specific Pretraining: In-Domain vs Out-of-Domain}
This study explores how the utilization of a model trained in a specific domain influences its performance when applied to a different domain. To conduct our experiments, we devised two primary approaches: in-domain pretraining and out-of-domain pretraining. In the case of in-domain pretraining, we initially trained a multilingual model on the MLS dataset. Subsequently, using this model as a seed, we further trained monolingual models on language-specific data exclusively from the MLS dataset. On the other hand, out-of-domain pretraining involved training a multilingual model on the in-house dataset and then using it as a seed to train monolingual models on language-specific data from the MLS dataset.

Our comprehensive experimentation aims to uncover the subtle dynamics of domain-specific pretraining, specifically delving into the distinctions between in-domain and out-of-domain pretraining. Our results reveal that in-domain pretraining led to a 36.2\% improvement compared to monolingual baselines. Conversely, out-of-domain pretraining demonstrated a more substantial impact with a 46.4\% Word Error Rate Reduction (WERR) against the baseline.

\begin{table}[h]
\begin{center}
\scalebox{0.78}{
\begin{tabular}{|cccc|}
\hline
\multicolumn{4}{|c|}{\textbf{In-domain pretraining | Seed: MLS Multilingual, Target: MLS}}                                           \\ \hline
\multicolumn{1}{|c|}{\textbf{Model}}            & \multicolumn{1}{c|}{\textbf{FR}} & \multicolumn{1}{c|}{\textbf{IT}} & \textbf{Avg} \\ \hline
\multicolumn{1}{|c|}{A. Monolingual ASR {\scriptsize{(baseline)}}}           & \multicolumn{1}{c|}{14.99}       & \multicolumn{1}{c|}{27.78}       & 21.38        \\ \hline
\multicolumn{1}{|c|}{Seeded monolingual MinWER ASR} & \multicolumn{1}{c|}{9.71}        & \multicolumn{1}{c|}{17.57}       & 13.64~\scriptsize{(36.2\%$\downarrow$})        \\ \hline
\multicolumn{4}{|c|}{\textbf{Out-of-domain pretraining | Seed: In-house Multilingual, Target: MLS}}                                  \\ \hline
\multicolumn{1}{|c|}{A. Monolingual ASR {\scriptsize{(baseline)}}}           & \multicolumn{1}{c|}{15.91}       & \multicolumn{1}{c|}{31.92}       & 23.93        \\ \hline
\multicolumn{1}{|c|}{Seeded monolingual MinWER ASR} & \multicolumn{1}{c|}{9.61}        & \multicolumn{1}{c|}{16.01}       & 12.81~\scriptsize{(46.4\%$\downarrow$})       \\ \hline
\end{tabular}
}
\end{center}
\caption{Impact of In-Domain vs Out-of-Domain Pretraining on Low-Resource Languages Compared to Multilingual and Monolingual Baselines.}
\end{table}

These findings reveals the effectiveness of employing out-of-domain pretraining, leading to an approximately 28.1\% higher WERR compared to in-domain pretraining. Leveraging a pretrained model trained on more diverse data of the same languages proved beneficial, enhancing model generalization and leading to significant performance improvements.


\subsection{Effect of Transfer Learning on Rare and Non-Rare Words}

This study aims to delve into the differntial effectiveness of transfer learning in enhancing both rare and non-rare words. In our experiment, words are categorized as rare or non-rare based on their frequency of occurrence. Specifically, rare words are identified as those with infrequent occurrences, typically falling below a predetermined threshold, while non-rare words are considered to be more commonly used.

Our investigation focused on the impact of in-domain and out-of-domain pretraining on the performance of rare and non-rare words. Interestingly, in the context of in-domain pretraining, non-rare words demonstrated a higher benefit compared to rare words, with rare words exhibiting a 32.1\% improvement, while non-rare words showed a 43.2\% WERR against the baseline. On the other hand, out-of-domain pretraining proved to be more effective for rare words than non-rare words, showcasing a 46.2\% WERR for rare words and a 42.1\% WERR for non-rare words. Considering out-of-domain pretraining as a more diverse approach contributes significantly to overall improvement. However, in the comparison between rare and non-rare words, it becomes evident that rare words experienced a greater benefit in this particular scenario.
\begin{table}[h]
\begin{center}
\scalebox{0.56}{
\begin{tabular}{|ccccccc|}
\hline
\multicolumn{7}{|c|}{\textbf{In-domain pretraining | Seed: MLS Multilingual, Target: MLS}}                                                                                                                                                           \\ \hline
\multicolumn{1}{|c|}{\multirow{2}{*}{\textbf{Model}}} & \multicolumn{3}{c|}{\textbf{Rare}}                                                                      & \multicolumn{3}{c|}{\textbf{Non-rare}}                                             \\ \cline{2-7} 
\multicolumn{1}{|c|}{}                                & \multicolumn{1}{c|}{\textbf{FR}} & \multicolumn{1}{c|}{\textbf{IT}} & \multicolumn{1}{c|}{\textbf{Avg}} & \multicolumn{1}{c|}{\textbf{FR}} & \multicolumn{1}{c|}{\textbf{IT}} & \textbf{Avg} \\ \hline
\multicolumn{1}{|c|}{A. Monolingual ASR}                 & \multicolumn{1}{c|}{32.33}       & \multicolumn{1}{c|}{56.22}       & \multicolumn{1}{c|}{44.26}        & \multicolumn{1}{c|}{13.58}       & \multicolumn{1}{c|}{27.11}       & 20.29        \\ \hline
\multicolumn{1}{|c|}{Seeded monolingual MinWER ASR}       & \multicolumn{1}{c|}{23.34}       & \multicolumn{1}{c|}{36.67}       & \multicolumn{1}{c|}{30.01~\scriptsize{(32.1\%$\downarrow$})}        & \multicolumn{1}{c|}{8.73}        & \multicolumn{1}{c|}{14.29}       & 11.51~\scriptsize{(43.2\%$\downarrow$})       \\ \hline
\multicolumn{7}{|c|}{\textbf{Out-of-domain pretraining | Seed: In-house Multilingual, Target: MLS}}                                      \\ \hline
\multicolumn{1}{|c|}{A. Monolingual ASR}                 & \multicolumn{1}{c|}{34.26}       & \multicolumn{1}{c|}{31.71}       & \multicolumn{1}{c|}{32.98}        & \multicolumn{1}{c|}{14.76}       & \multicolumn{1}{c|}{25.77}       & 20.23        \\ \hline
\multicolumn{1}{|c|}{Seeded monolingual MinWER ASR}       & \multicolumn{1}{c|}{20.16}       & \multicolumn{1}{c|}{15.32}       & \multicolumn{1}{c|}{17.74~\scriptsize{(46.2\%$\downarrow$})}        & \multicolumn{1}{c|}{7.51}        & \multicolumn{1}{c|}{15.91}       & 11.71~\scriptsize{(42.1\%$\downarrow$})        \\ \hline
\end{tabular}
}
\end{center}
\caption{Influence of In-Domain vs Out-of-Domain Pretraining on Rare and Non-Rare Words.}
\end{table}
\section{Discussion}
We further conducted multiple ablation studies to comprehend the impact of hold and warm-up steps in model training, evaluate the zero-shot language performance with our pretraining approach, and analyze how transfer learning influences the model convergence. These aspects will be discussed in the following subsections.
\\
\subsection{Influence of Warm-Up and Hold Steps}
In non-transfer learning ASR model training, we opted for a fixed 3-step learning rate (LR) \cite{park2019specaugment} during warm-up and hold steps, aiming to balance rapid early convergence with stable later-stage training for optimized performance. However, during the transfer learning stage, we observed that setting Warm-Up and Hold Steps to $0$ expedites model convergence and reduces training time by eliminating initial steps that may hinder the adaptation process. Warm-Up Steps typically involve a gradual adjustment of learning rates, allowing the model to acclimate to the new task. However, in scenarios where the pretrained model is already well-established, starting with a warm-up may be unnecessary. Similarly, Hold Steps, which involve maintaining a constant learning rate, may not be essential when transferring knowledge from a pretrained model with well-tuned parameters. By bypassing these initial adjustments, the model can promptly leverage its existing knowledge and adapt swiftly to the new task, resulting in faster convergence and more efficient training. 

\subsection{Effect of pretraining on zero-shot language performance}
In the follow-up study, we investigated the impact of transfer learning on the performance of zero-shot languages, referring to languages not included in the pretraining model. Our approach involved training a multilingual model using datasets in English, Italian, and French languages. Subsequently, we utilized this model as a seed to initiate the training of monolingual models specifically for German and Spanish. Notably, our findings highlight a significant enhancement in the performance of the Spanish model, achieving a remarkable 26.6\% WERR. Conversely, the German model exhibited a WER regression of 44.8\%. It is important to note that both models were evaluated against baselines without any pretraining.

\begin{table}[h]
\begin{center}
\scalebox{0.8}{

\begin{tabular}{|c|c|c|}
\hline
\textbf{Model}                & \textbf{ES} & \textbf{DE} \\ \hline
Monolingual ASR               & 8.96        & 12.23       \\ \hline
Seeded monolingual MinWER ASR & 6.58~\scriptsize{(26.6\%$\downarrow$})       & 17.71~\scriptsize{(44.8\%$\uparrow$})       \\ \hline
\end{tabular}
}
\end{center}
\caption{Illustrating the impact of pretraining on zero-shot languages: exploring scenarios where the language is not part of the pretraining stage.}
\end{table}

We speculate that the reason Spanish works well with English, Italian, and French is because they share similar language roots. On the other hand, German might not fit as well due to its potential linguistic differences and distinct acoustic patterns. This suggests that when creating a pretraining model, combining languages with similar linguistic backgrounds could significantly enhance its performance with languages that were not part of the initial training set.

\subsection{Impact of Transfer Learning on Model Convergence}
In our investigation, initializing the model with a fully converged pretrained seed model yielded noteworthy benefits in terms of faster convergence compared to training the model from scratch. On average, our findings indicate a substantial reduction in training time, with the transfer learning model demonstrating a remarkable decrease from approximately 7 days (168 hours) to around 2 days (48 hours). Additionally, this accelerated convergence is accompanied by a significant reduction in power consumption, dropping from 1196 kWH to 299 kWH, marking a 75\% decrease. This study sheds light on the efficiency gains achieved through the incorporation of transfer learning in the model initialization process.

The faster convergence of the transfer learning model is due to leveraging knowledge from the fully converged pretrained seed model. By utilizing learned features, representations, and patterns, the transfer learning model benefits from a more informed initialization, facilitating rapid adaptation to the target task specifics. This results in fewer iterations needed for parameter optimization and achieving convergence. The knowledge transfer from the pretrained model expedites the learning process.

\section{Conclusion}
In summary, the study highlights the consistent effectiveness of incorporating multilingual pretraining at the baseline stage for enhancing monolingual ASR, particularly in languages like Italian and French. Notably, out-of-domain pretraining proves even more advantageous compared to in-domain pretraining across both baseline scenarios. Additionally, the research underscores the positive impact of multilingual pretraining on both rare and non-rare words, with non-rare words showing a slightly greater influence in in-domain pretraining, while rare words benefit more from out-of-domain pretraining. These findings collectively underscore the strategic importance of pretraining approaches in optimizing ASR models for enhanced language-specific and cross-linguistic performance.

In future work, we intend to explore targeted layer-wise transfer learning, focusing on key layers for effective knowledge transfer rather than transferring the entire encoder-based pretraining. We also aim to use language-based identification to selectively transfer knowledge within the same language family, addressing cases where language relationships may impact performance differently, such as observed effects on German and Spanish. These steps will enhance the precision of multilingual transfer learning.



\bibliographystyle{IEEEtran}
\bibliography{mybib}

\end{document}